\documentclass{article}    

\usepackage{graphicx}          

\usepackage{amsmath}
\usepackage{amsfonts}
\usepackage{cite}
\usepackage{fancyhdr}

\usepackage{hyperref}
\usepackage{multirow}
\usepackage[ansinew]{inputenc}

\hypersetup{
    colorlinks=true,       
    linkcolor=blue,          
    citecolor=blue,        
    urlcolor=blue          
}

\newcommand{\dpl}{\langle}
\newcommand{\dpm}{,}
\newcommand{\dpr}{\rangle}
\newcommand{\dotprod}[2]{\dpl #1 \dpm #2 \dpr}

\newcommand{\st}{\mathrm{subject~to}}

\newcommand{\eye}{\mathbf{I}}

\newcommand{\Kx}{\mathbf{K}}
\newcommand{\Ky}{\mathbf{L}}

\newtheorem{lemma}{Lemma}[section]

\begin{document}

\title{Learning Output Kernels for Multi-Task Problems}

\author{Francesco Dinuzzo \thanks{Max Planck Institute for Intelligent Systems, Spemannstrasse 38,72076 Tübingen, Germany. E-mail: fdinuzzo@tuebingen.mpg.de}}    
\date{}

\maketitle

\begin{abstract}                          
Simultaneously solving multiple related learning tasks is beneficial under a variety of circumstances, but the prior knowledge necessary to correctly model task relationships is rarely available in practice. In this paper, we develop a novel kernel-based multi-task learning technique that automatically reveals structural inter-task relationships. Building over the framework of output kernel learning (OKL), we introduce a method that jointly learns multiple functions and a low-rank multi-task kernel by solving a non-convex regularization problem. Optimization is carried out via a block coordinate descent strategy, where each subproblem is solved using suitable conjugate gradient (CG) type iterative methods for linear operator equations. The effectiveness of the proposed approach is demonstrated on pharmacological and collaborative filtering data.
\end{abstract}

\section{Introduction} \label{sec01}

Combining multiple datasets for solving related inference problems is a common and successful practice in several domains such as econometrics and marketing analysis, recommender systems on the web, bioinformatics, pharmacology, and many others. Indeed, simultaneously solving multiple inference problems can improve performances, provided that the relationships between them are correctly modeled. The themes of transfer learning, multi-task learning or “learning to learn” \cite{Caruana97, Thrun98, Heskes00} have attracted considerable attention in the literature, see \cite{Pan10} for a recent survey. A possible way to enforce relationships between the tasks is to extract a set of shared features. This can be done by employing multi-layer neural networks where the hidden layer is shared among the tasks \cite{Caruana97}, or also within a convex regularization framework \cite{Argyriou08}. Task relationships can be also modeled within the framework of Gaussian Processes estimation \cite{Schwaighofer05, Yu05, Bonilla08, Pillonetto10, Wang10, Birlitiu10} by designing a joint covariance function. A variety of other models have been proposed to represent and exploit inter-task relationships \cite{Ando05, Yu07,  Xue07,  Zhang08, Qi08, Kato10}.


Correctly modeling the inter-task relationships for a given application is critical but not always easy. In fact, in many inference problems, the tasks are known to be related with each other, but the available prior knowledge is not sufficient to model the relationships in advance. For this reason, some recent works have been focusing on inferring inter-task relationships automatically from the data, while solving the multi-task learning problem. One possibility is to assume that the tasks can be clustered into homogeneous groups and try to learn such clustering from the data \cite{Bakker03}. Optimization-based approaches that jointly infer the task parameters and the inter-task relationships in the form of a similarity matrix include the spectral regularization approach of \cite{Argyriou07} and the method of \cite{Zhang10} based on convex optimization. An online method has been also proposed recently to learn multiple linear classifiers as well as a task relationship matrix \cite{Saha11}.

Regularized kernel methods \cite{Scholkopf01b} have been employed successfully in a variety of single-task learning problems. Their extension to the multi-task setting \cite{Evgeniou05} calls for the design of suitable operator-valued kernels that model similarities of both inputs and tasks. Some classes of operator-valued kernels have been recently reviewed in \cite{Alvarez12}. Within the framework of regularization in RKHS of vector-valued functions, the problem of inferring task relationships boils down to the problem of learning a multi-task kernel. A possible way to address this problem consists in learning a linear combination of task-specific kernels \cite{Widmer10}, within the framework of multiple kernel learning (MKL). Recently, a class of output kernel learning (OKL) methods \cite{Dinuzzo11, Dinuzzo11c} has been introduced in the context of multi-output learning problems (such as vectorial regression, multi-class and multi-label classification) to automatically synthesize a decomposable matrix-valued kernel that encodes the relationships between the output components. Differently from MKL, OKL methods don't require the specification of a set of basis kernels to be combined, as the output kernel is searched into the whole cone of positive semidefinite kernels.

This paper explores the possibility of automatically learning the relationships between multiple tasks via output kernel learning. To this end, we extend a technique for learning low-rank output kernels proposed in \cite{Dinuzzo11b} to the multi-task setting. The low-rank encouraging regularization can be shown to be useful both for computational reasons and learning performances. The extension developed in this paper is based on the use of a suitable weighted loss that allows for multiple datasets with different input sampling patterns. The resulting method can be also interpreted as a non-linear kernel based generalization of low-rank matrix completion techniques. Even in the simpler setup of linear low-rank matrix factorization, the case of incomplete observations has been recognized to be computationally challenging, see e.g. \cite{Srebro03}, since existing optimization techniques based on eigendecompositions do not carry over to the weighted case. For this reason, we have developed a new strategy to solve the optimization problem. Similarly to \cite{Zhang10}, we learn multiple tasks as well as their relationships by solving a joint optimization problem. However, differently from \cite{Zhang10} and \cite{Saha11}, that learn multiple linear functions by convex optimization, we learn multiple non-linear functions by solving a non-convex optimization problem. Albeit non-convex, the problem exhibits a special structure that allows for effective optimization via a block-coordinate descent procedure based on the iterative application of the conjugate gradient (CG) algorithm for linear operator equations. Our method is also related to the technique derived in \cite{Bonilla08} in the context of Bayesian estimation of Gaussian Processes, that allows to learn a similarity (covariance) matrix between the tasks. While \cite{Bonilla08} aims at optimizing a marginal likelihood type functional, our method is based on the minimization of a functional with trace norm regularization plus rank constraint. In \cite{Bonilla08}, the authors adopt a general purpose gradient-descent solver to optimize their objective functional, whereas in this paper we develop a novel optimization strategy that is specifically designed to solve the proposed OKL optimization problem. Differently from \cite{Bonilla08}, our method is able to deal with the case of incomplete sampling, and automatically encourages low-rank solutions without introducing relaxations of the original problem.

\section{Weighted output kernel learning}\label{sec02}

Let $\eye$ denote the identity matrix and $e$ the vector of all ones (of suitable dimensions). For any matrix $\mathbf{A}$, let $\mathbf{A}^T$ denote the transpose, $\textrm{tr}(\mathbf{A})$ the trace, $\textrm{rank}(\mathbf{A})$ the rank, $\textrm{vec}(\mathbf{A})$ the vectorization operator, $\mathbf{A}^{\dag}$ the pseudo-inverse, $\| \mathbf{A} \|_F = \left(\textrm{tr}\left(\mathbf{A}^T \mathbf{A}\right)\right)^{1/2}$ the Frobenius norm, and $\| \mathbf{A} \|_{*} := \textrm{tr}\left(\left(\mathbf{A}^T\mathbf{A}\right)^{1/2}\right)$ the nuclear norm. For any pair of matrices of the same size $\mathbf{A}, \mathbf{B}$, let $\langle \mathbf{A}, \mathbf{B} \rangle_{F} := \textrm{tr}(\mathbf{A}^T \mathbf{B})$ denote the Frobenius inner product. The symbols $\otimes$ and $\odot$ denote the Kronecker product and the Hadamard (element-wise) product, respectively.  Finally, let $\mathbb{S}^m_+$ denote the closed cone of positive semidefinite matrices of order $m$, and
\[
\mathbb{S}^{m,p}_+ = \left\{\mathbf{A} \in \mathbb{S}^m_+: \textrm{rank}(\mathbf{A}) \leq p \right\} \subseteq \mathbb{S}^m_+
\]
\noindent the cone of positive semidefinite matrices with rank less than or equal to $p$.

\subsection{Output kernel learning for multi-task problems} \label{sec02-01}

Consider the problem of learning several functions (tasks) $g_j:\mathcal{X} \rightarrow \mathbb{R}$ ($j=1,\ldots,m$) from multiple datasets of input-output pairs $(x_{ij}, y_{ij})$. Since the input set $\mathcal{X}$ is common to all the tasks, the functions $g_j$ can be combined into a single vector-valued function $g:\mathcal{X} \rightarrow \mathbb{R}^m$, to be learned from a single dataset of $\ell$ pairs $(x_{i},y_{i})$, where the output data $y_i$ are now vectors that may contain missing components. Let $\mathbf{W} \in \left\{0,1\right\}^{\ell \times m}$ denote a binary \emph{weight matrix} specifying which output components are missing for each example. More precisely, the $i$-th row of the weight matrix is a binary vector $w_i$ with zeros in correspondence with the missing components of $y_i$. Since the weight matrix is always available, we can assume without loss of generality that all the missing output components have been imputed with zeros.

In the following, we describe a regularization approach where the function $g$ is searched into a reproducing kernel Hilbert space (RKHS) $\mathcal{H}$ of vector-valued functions $g:\mathcal{X} \rightarrow \mathbb{R}^m$ associated with a decomposable reproducing kernel $H$ of the form
\[
H(x_1,x_2) = K_X(x_1,x_2) \cdot \Ky,
\]
where $K_X$ is positive semidefinite scalar kernel (called \emph{input kernel}) and $\Ky \in \mathbb{S}^m_+$ is a symmetric positive semidefinite matrix (called \emph{output kernel}). For more details about RKHS of vector-valued functions, see \cite{Micchelli05}.

The function $g \in \mathcal{H}$ and the output kernel $\Ky$ are simultaneously learned by solving a joint regularization problem of the form
\begin{equation}
\label{EQ01}
\min_{\substack{\Ky \in \mathbb{S}^{m,p}_+\\ g \in\mathcal{H}}}\left(\sum_{i=1}^{\ell}\frac{\left\|w_i \odot (g(x_{i})-y_{i})\right\|_2^2}{2\lambda} + \frac{\|g \|_{\mathcal{H}}^2}{2}  + \frac{\textrm{tr}(\Ky)}{2}\right).
\end{equation}
The objective functional contains a data-fitting term, taking into account the prediction error in correspondence with the observed output components, a regularization term on the unknown function $g$, and a trace regularization on the output kernel. Optimization of the output kernel is carried over the low-rank cone $\mathbb{S}^{m,p}_+$, thus also imposing a hard rank constraint.

In view of the representer theorem \cite{Kimeldorf71, Dinuzzo12}, the minimization problem with respect to $g$ admits a solution of the form
\[
 g^*(x) =  \Ky\sum_{i=1}^{\ell}c_i K_X(x,x_i),
\]
with suitable coefficient vectors $c_i \in \mathbb{R}^m$. Letting $\Kx \in \mathbb{S}_+^{\ell}$ be such that $\Kx_{ij} = K(x_i,x_j)$, the following finite-dimensional optimization problem is obtained

\begin{equation}
  \label{EQ02}
  \min_{\substack{\Ky \in \mathbb{S}^{m,p}_+ \\ \mathbf{C} \in \mathbb{R}^{\ell \times m}}} \left(\frac{\|\mathbf{Y}-\Kx\mathbf{C}\Ky\|_{\mathbf{W}}^2}{2\lambda}+ \frac{\dotprod{\mathbf{C}^T\Kx\mathbf{C}}{\Ky}_{F}}{2}+ \frac{\textrm{tr}(\Ky)}{2}\right),
\end{equation}
\noindent where the matrices $\mathbf{Y}, \mathbf{C} \in \mathbb{R}^{\ell \times m}$ are defined as
\[
\mathbf{Y} = \left(y_1, \ldots, y_{\ell}\right)^T,
\qquad
\mathbf{C} = \left(c_1, \ldots, c_{\ell}\right)^T,
\]
\noindent and the (semi)-norm $\| \cdot \|_{\mathbf{W}}$ is given by
\[
\|\mathbf{A}\|_{\mathbf{W}} = \|\mathbf{W} \odot \mathbf{A}\|_F.
\]
\noindent Observe that $\mathbf{Y}$ is a sparse matrix (missing values have been imputed with zeros). Moreover, since $\mathbf{W}$ is a binary matrix with the same sparsity pattern of $\mathbf{Y}$, we also have
\begin{equation}\label{EQ03}
\mathbf{Y} = \mathbf{W} \odot \mathbf{Y}.
\end{equation}

Within the formulation of problems (\ref{EQ01}) and (\ref{EQ02}), a first possibility is to simply set $p = m$. With such a choice, the hard constraint on the rank is not present, but the trace regularization will still encourage low-rank solutions, see e.g. \cite{Fazel01}. It is nevertheless convenient to allow for the more general choice $p \le m$. A first reason is that, by providing an explicit bound $p < m$ on the rank, it is possible to control the computational and memory requirements of the method before running the optimization. This holds since, within the optimization framework developed in the next section, only matrices of rank $p$ are stored and manipulated, whereas the full matrix $\Ky$ is never formed explicitly. A second reason is that, when the regularization parameter $\lambda$ is small enough, the hard rank constraint becomes active, and may yield interesting solutions that are not obtainable by using only the trace regularization (see, for example, the experiment in subsection \ref{sec04-01}).

The objective functional of (\ref{EQ02}) is not jointly (quasi)-convex. However, it is convex separately with respect to both $\Ky$ and $\mathbf{C}$. In addition, for $p=m$ it is an invex function \cite{Mishra08} in the interior of the feasible set, meaning that every stationary point is a global minimizer. When all the output components are observed, i.e. $\mathbf{W} = e e^T$, problem (\ref{EQ02}) can be attacked using techniques based on eigendecompositions. Since these techniques do not apply anymore for general weight matrices, in the following we develop a new strategy to obtain a minimizer of (\ref{EQ02}).

\subsection{Equivalent formulations} \label{sec02-03}

Before going into the details of the optimization procedure proposed to solve (\ref{EQ02}), is it useful to introduce some equivalent reformulations of the optimization problem. The proofs of the two following Lemmas are reported in the appendix.

\begin{lemma}\label{LEM01}
If $\left(\mathbf{A}, \mathbf{B}\right)$ is an optimal solution of the following problem:
\begin{equation}\label{EQ04}
\min_{\substack{\mathbf{A} \in \mathbb{R}^{\ell \times p}\\\mathbf{B} \in \mathbb{R}^{m \times p}}} \left(\frac{\|\mathbf{Y}-\Kx\mathbf{A}\mathbf{B}^T\|_{\mathbf{W}}^2}{2\lambda}+ \frac{\dotprod{\mathbf{A}}{\Kx\mathbf{A}}_{F}}{2}+ \frac{\|\mathbf{B}\|_F^2}{2}\right),
\end{equation}
\noindent then any pair $\left(\mathbf{C}, \Ky\right)$ such that
\[
\Ky = \mathbf{B}\mathbf{B}^T, \qquad \mathbf{A}=\mathbf{C}\mathbf{B},
\]
is an optimal solution of problem (\ref{EQ02}).
\end{lemma}

Lemma \ref{LEM01} provides a new formulation of the low-rank output kernel learning problem (\ref{EQ02}) that involves only ``thin'' matrices $\mathbf{A}$ and $\mathbf{B}$, whose size can be controlled by selecting the parameter $p$. Such formulation turns out to be particularly convenient for optimization purposes. It is also insightful to observe that problem (\ref{EQ02}) can be reformulated as a reduced rank least square problem with a ``kernelized'' nuclear norm regularization.

\begin{lemma}\label{LEM02}
If $\mathbf{\Theta}$ is an optimal solution of the following problem:
\begin{align*}
\min_{\mathbf{\Theta} \in \mathbb{R}^{\ell \times m}}& \left[\frac{\|\mathbf{Y}-\Kx\mathbf{\Theta}\|_{\mathbf{W}}^2}{2\lambda}+ \textup{tr}\left(\left(\mathbf{\Theta}^T\Kx\mathbf{\Theta}\right)^{1/2}\right)\right],\\
& \st \quad \textup{rank}(\mathbf{\Theta}) \leq p,
\end{align*}
\noindent then the pair
\[
\Ky = \left(\mathbf{\Theta}^T\Kx\mathbf{\Theta}\right)^{1/2}, \quad \mathbf{C} = \Ky^{\dag} \mathbf{\Theta},
\]
\noindent is an optimal solution of problem (\ref{EQ02}).
\end{lemma}

Lemma \ref{LEM02} shows that the low-rank output kernel learning problem (\ref{EQ02}) is a non-linear kernelized generalization of least squares low-rank matrix approximation (RMF) with nuclear norm regularization. Indeed, RMF can be obtained as a particular case by choosing the input kernel as a Kronecker delta kernel
\[
K(x_1,x_2) = \delta_K(x_1,x_2) =\left\{
                                  \begin{array}{ll}
                                    1, & x_1 = x_2 \\
                                    0, & \hbox{else}
                                  \end{array}
                                \right.
 \]
so that
\[
\Kx = \mathbf{I}, \qquad \textup{tr}\left(\left(\mathbf{\Theta}^T\Kx\mathbf{\Theta}\right)^{1/2}\right) = \|\mathbf{\Theta}\|_*.
\]
The related MMMF (maximum-margin matrix factorization) technique \cite{Srebro05, Rennie05} would also correspond to the case $\Kx = \mathbf{I}$, but with hinge-type (SVM) losses instead of the square loss.

\section{A block coordinate descent strategy} \label{sec03}

In the following, we focus on the solution of (\ref{EQ04}). If $p$ is much smaller than $m$, handling the low-dimensional factor $\mathbf{B}$ is much more convenient than directly handling the full output kernel matrix. Let $J(\mathbf{A},\mathbf{B})$ denote the objective functional of (\ref{EQ04}). Observe that, although $J$ is non-convex, it is unconstrained and separately convex with respect to both factors. Therefore, it is natural to adopt a block coordinate descent technique that iteratively alternates between optimization with respect to the two blocks. Clearly, other optimization strategies are also possible, but the block coordinate descent approach is memory efficient, robust, and simple to implement. In addition, it turns out to behave well in practice, since very few iterations are typically sufficient to obtain a good solution, especially when combined with a warm-start regularization path procedure (see subsection \ref{sec03-03}). As shown in the following, the two subproblems boil down to the solution of linear equations of the form
\begin{eqnarray}\label{EQ05}
\mathcal{L}_A\mathbf{A} & = & y_A,\\
\label{EQ06}
\mathcal{L}_B\mathbf{B} & = & y_B
\end{eqnarray}
\noindent where $\mathcal{L}_A$ and $\mathcal{L}_B$ are linear operators mapping matrices into vectors. In the following subsections, we derive equations (\ref{EQ05})-(\ref{EQ06}), and discuss the application of suitable iterative methods for solving them.

\subsection{Sub-problem w.r.t. $\mathbf{A}$} \label{sec03-01}

The sub-problem w.r.t. $\mathbf{A}$ is the most numerically challenging of the two since, in general, the linear operator $\mathcal{L}_A$ is not symmetric. For any fixed $\mathbf{B}$, a necessary and sufficient condition for $\mathbf{A}$ to be optimal is obtained by setting to zero the partial derivative of the objective functional:
\[
\frac{\partial J}{ \partial \mathbf{A}}(\mathbf{A},\mathbf{B}) = \Kx \left[\frac{\mathbf{W} \odot \left(\Kx\mathbf{A}\mathbf{B}^T- \mathbf{Y}\right)\mathbf{B}}{\lambda} + \mathbf{A}\right] = \mathbf{0}.
\]
\noindent A sufficient condition is given by
\[
\mathbf{W} \odot \left(\Kx\mathbf{A}\mathbf{B}^T\right)\mathbf{B} + \lambda \mathbf{A} = \mathbf{Y}\mathbf{B},
\]
\noindent where we have used (\ref{EQ03}) in order to simplify the right hand side. If the weight matrix is full, this last equation reduces to a discrete-time Sylvester equation, a well studied class of linear matrix equations, see e.g.  \cite{Sima96}. However, for general $\mathbf{W}$, the optimality condition is not a Sylvester equation anymore, due to presence of the Hadamard product. Now, letting
\[
\mathbf{W}_D = \textrm{diag}(\textrm{vec}(\mathbf{W})),
\]
\[
\mathcal{L}_A\mathbf{A} = \left[\left(\mathbf{B}^T\otimes \mathbf{I}\right) \mathbf{W}_D  \left(\mathbf{B}\otimes \Kx\right) + \lambda \mathbf{I}\right]  \textrm{vec}(\mathbf{A}),
\]
\[
y_B = \textrm{vec}(\mathbf{Y}\mathbf{B}),
\]
\noindent and using the matrix identities (\ref{EQ08}) and (\ref{EQ10}), the sufficient optimality condition can be rewritten in the form (\ref{EQ05}).

If $\mathbf{W}$ is full, it is easy to verify, using the mixed-product identity (\ref{EQ09}) for the Hadamard product, that the operator $\mathcal{L}_A$ reduces to
\[
\mathcal{L}_A \mathbf{A}= \left[\left(\mathbf{B}^T\mathbf{B}\right)\otimes \Kx + \lambda \mathbf{I}\right]\textrm{vec}(\mathbf{A}),
\]
\noindent and is therefore symmetric and positive definite. In such a case, one can either apply the conjugate gradient algorithm, or also use a procedure based on eigendecompositions in order to solve (\ref{EQ05}). Unfortunately, these methods cannot be applied for general weight matrices, since the operator $\mathcal{L}_A$ is not guaranteed to be symmetric.

A first possible way to attack the problem is trying to directly obtain a solution of the non-symmetric operator equation (\ref{EQ05}) by means of iterative methods that can handle non-symmetric equations, such as the generalized minimal residual method (GMRES) \cite{Saad86}. However, the computational requirements of GMRES grow with the number of iterations and, for large scale problems, a restart strategy is needed to limit the memory requirements at the expense of convergence speed. The alternative is to introduce a change of variable to make the problem symmetric, and then apply a preconditioned conjugate gradient (CG) algorithm \cite{Hestenes52} on the new linear equation. Generally, CG requires less computational resources than GMRES, since the search directions are obtained via simple recursive updates and a restart strategy is not required. For additional details about these and others iterative method for solving linear equations, see e.g. \cite{Saad03}.

Another way to address non-symmetry is to look for a change of variable that makes the problem symmetric. A first possible change of variable derives from the observation that the optimal solution must be in the form $\mathbf{A} = \mathbf{C}\mathbf{B}$. By using this representation, one can equivalently solve a new linear equation with respect to $\mathbf{C}$:
\[
\mathbf{W} \odot \left[\Kx \left(\mathbf{W} \odot \mathbf{C}\right)\mathbf{B}\mathbf{B}^T\right]+\lambda \mathbf{C} = \mathbf{Y}.
\]
\noindent Once again, the equation can be rewritten in the form
\[
\mathcal{L}_C\mathbf{C} = \textrm{vec}(\mathbf{Y}),
\]
\noindent where
\[
\mathcal{L}_C\mathbf{C} = \left[\mathbf{W}_D \left((\mathbf{B}\mathbf{B}^T)\otimes\mathbf{K}\right)\mathbf{W}_D+\lambda\mathbf{I}\right]\textrm{vec}(\mathbf{C}).
\]
\noindent This time, the operator $\mathcal{L}_C$ is symmetric and positive definite, so that CG applies. A disadvantage of this approach is that, for $p \ll m$,  $\mathbf{C}$ is much higher-dimensional than $\mathbf{A}$. Nevertheless, when the weight matrix $\mathbf{W}$ is sparse, $\mathbf{C}$ is also a sparse matrix with the same sparsity pattern, and one can take advantage of this property.

Another approach is based on a factorization of the input kernel matrix of the form
\[
\mathbf{K} = \mathbf{F}\mathbf{F}^T.
\]
\noindent The factor $\mathbf{F}$ can be a Cholesky factor, a matrix square root, or also a low-rank factor. We can use the factorization to introduce a new change of variable
\begin{equation}\label{EQ07}
\mathbf{A}_F = \mathbf{F}^T \mathbf{A},
\end{equation}
\noindent so that the matrix $\mathbf{A}_F$ solves the equation
\[
\mathbf{F}^T\mathbf{W} \odot \left(\mathbf{F}\mathbf{A}_F\mathbf{B}^T\right)\mathbf{B} + \lambda \mathbf{A}_F = \mathbf{F}^T\mathbf{Y}\mathbf{B}.
\]
\noindent This last equation can be rewritten in vectorized form
\[
\mathcal{L}_F\mathbf{A}_F = \textrm{vec}(\mathbf{F}^T\mathbf{Y}\mathbf{B}),
\]
\noindent where
\[
\mathcal{L}_F\mathbf{A}_F  = \left[\left(\mathbf{B}^T\otimes \mathbf{F}^T\right) \mathbf{W}_D  \left(\mathbf{B}\otimes \mathbf{F}\right) + \lambda \mathbf{I}\right]  \textrm{vec}(\mathbf{A}_F).
\]

\noindent Independently of which factorization of $\Kx$ has been adopted, the linear operator $\mathcal{L}_F$ is symmetric and positive definite, so that CG can be applied. Once a solution $\mathbf{A}_F$ has been obtained, the matrix $\mathbf{A}$ can be recovered as a solution of the linear system (\ref{EQ07}).

In our experiments, all of the aforementioned approaches have been tested. The last approach based on the Cholesky factorization of the input kernel matrix turned out to be generally faster than the others, therefore we selected it for our current implementation.

\subsection{Sub-problem w.r.t. $\mathbf{B}$} \label{sec03-02}

The sub-problem w.r.t $\mathbf{B}$ is essentially a multiple ridge regression problem. First of all, observe that the objective functional depends on $\mathbf{A}$ only through the product $\mathbf{E} = \Kx \mathbf{A}$. Therefore, when $\mathbf{A}$ is fixed, a necessary and sufficient condition for $\mathbf{B}$ to be optimal is
\[
\frac{\partial J}{ \partial \mathbf{B}}(\mathbf{A},\mathbf{B}) = \mathbf{E}^T \frac{\mathbf{W} \odot \left(\mathbf{E} \mathbf{B}^T-\mathbf{Y}\right)}{\lambda} + \mathbf{B}^T = \mathbf{0}.
\]

\noindent In this case, it turns out that an expression for the rows of $\mathbf{B}$ can be even written in closed form. Indeed, let $b_j$ ($j=1, \ldots, m$) denote the rows of $\mathbf{B}$, $y^j$ and $w^j$ denote the columns of $\mathbf{Y}$ and $\mathbf{W}$, respectively. Then, we have
\[
b_j = \left(\mathbf{E}^T \textrm{diag}(w^j)\mathbf{E} + \lambda \mathbf{I}\right)^{-1}\mathbf{E} \textrm{diag}(w^j) y^j.
\]
Observe that the updates for the different rows can be computed in parallel. Finally, by letting
\[
\mathcal{L}_B\mathbf{B} = \left[(\mathbf{E}^T\otimes\mathbf{I})\textrm{diag}(\textrm{vec}(\mathbf{W}^T))(\mathbf{E}\otimes\mathbf{I})+\lambda\mathbf{I}\right] \textrm{vec}(\mathbf{B}),
\]
\noindent and
\[
y_B = \textrm{vec}(\mathbf{Y}^T\mathbf{E}),
\]
\noindent the optimality condition can be also rewritten in the form (\ref{EQ06}). Here, the operator $\mathcal{L}_B$ can be seen to be symmetric and positive definite, therefore it is possible to use a conjugate gradient method to obtain a solution.

\subsection{Implementation details} \label{sec03-03}

Several important details must be taken into account in order to guarantee a correct and efficient convergence behavior.

\subsubsection{Warm-start path procedure}

It is typically necessary to train the kernel machine for several different values of the regularization parameter $\lambda$. Generally, the regularization problem is better conditioned for large values of $\lambda$. On the same time, the solution is expected to depend continuously on the regularization parameter. For these reasons, it is convenient to initialize the optimization of each problem with the solution obtained with the previous value of $\lambda$, after sorting the different values of the regularization parameter in decreasing order. Such \emph{warm-start} regularization path strategy is very effective in practice: if the grid of values of $\lambda$ is sufficiently fine, one or two iterations of block coordinate descent for each value of $\lambda$ are often sufficient to converge with a good accuracy.

\subsubsection{Initialization}

It is possible to show that the rank of $\mathbf{B}$ cannot increase between an iteration of block-coordinate descent and the next. For this reason, at the very beginning of the warm-start procedure, $\mathbf{B}$ is initialized to a full-rank matrix. Otherwise, an optimal solution of high rank may be missed by the algorithm. Another issue comes from the fact that the origin is a stationary point of the objective functional for any value of $\lambda$. Although, for very large values of the regularization parameter, the origin is actually a global minimizer, this is not the case anymore for small values. Now, in view of the warm-start procedure, the algorithm may never move away from the origin if the initial $\lambda$ is very large. To safeguard against this behavior, $\mathbf{B}$ is re-initialized to a full-rank matrix every time it becomes too small (for example, w.r.t. the Frobenius norm).

\section{Experiments} \label{sec04}

In this section, we analyze the performance of weighted OKL on a variety of multi-task problems, including pharmacokinetic-pharmacodynamics (PK-PD) problems, and popular collaborative filtering benchmarks (MovieLens datasets). In all the experiments on real data, the performance of weighted OKL is compared with the following methods:
\begin{itemize}
  \item \textbf{Independent single-task learning}: corresponds to fixing the output kernel as $\Ky = \mathbf{I}$ instead of optimizing it, thus learning each task independently of the others.
  \item \textbf{Pooled single-task learning}: corresponds to fixing the output kernel as $\Ky = e e^T$, thus assuming that all the tasks are the same.
  \item \textbf{Regularized Matrix Factorization} (RMF): corresponds to fixing the input kernel as the Kronecker delta kernel, so that $\Kx = \mathbf{I}$.
\end{itemize}
All the experiments have been run in a MATLAB environment with an Intel i5 CPU 2.4 GHz, 4 GB RAM.

\subsection{Reconstruction and denoising of multiple signals} \label{sec04-01}

\begin{figure}[h]
  \centering
  \includegraphics[width=1\textwidth]{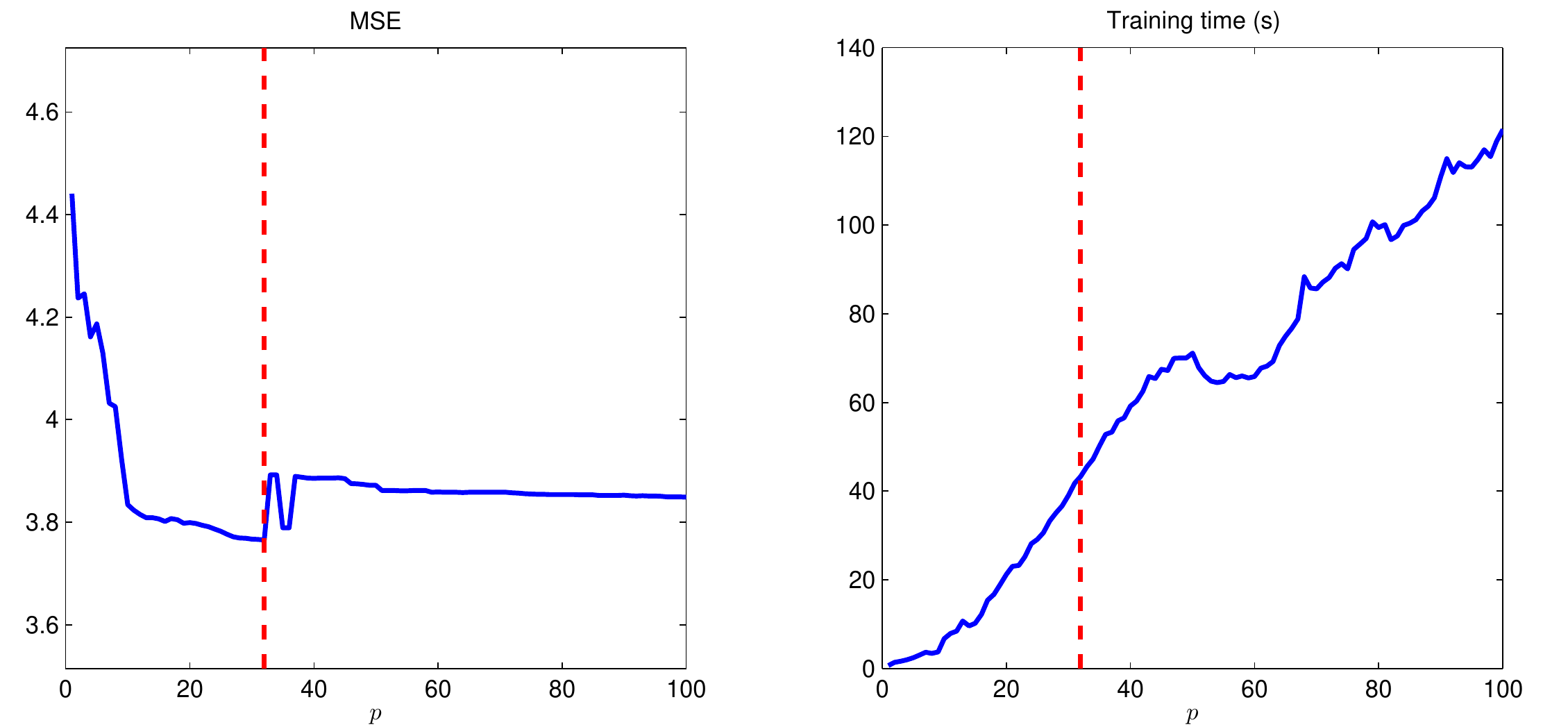}
  \caption{The continuous lines represent the MSE (left panel) and training time (right panel) of OKL, as a function of the parameter $p$. The vertical dashed line corresponds to the value of $p$ that minimizes MSE.}
  \label{fig:LowRankGP}
\end{figure}

In order to analyze the dependence of the computation time on the rank bound parameter $p$ of the proposed OKL method, we conducted some controlled experiments on synthetic data.  The experiments show both the computational and the potential predictive advantage induced by the hard rank constraint. First of all, we generated 50 independent realizations $Z_k$, $(k = 1, \ldots, 50)$ of a Gaussian Process on the interval $[-1, 1]$ of the real line with zero-mean and covariance function
\[
\textrm{cov}\left[(Z_k)_{x_1},(Z_k)_{x_2}\right] = \textrm{exp}(-10|x_1-x_2|).
\]
\noindent Then, we generated $m=200$ new processes $U_j$ as
\[
U_j = \sum_{k=1}^{50} B_{jk} Z_k,
\]
\noindent where the mixing coefficients $B_{jk}$ are independently drawn from a uniform distribution on the interval $[0, 1]$. Output data $y_{ij}$ have been generated by sampling the processes $U_j$ in correspondence with 100 points in the interval $[-1, 1]$ randomly drawn from a uniform grid of 200 points, and corrupting them by adding a zero-mean Gaussian noise with a signal to noise ratio of 1:1. For each process $U_j$, only 70 out of the 100 inputs have been used for training, while the remaining 30 are used for tuning the regularization parameter $\lambda$.

The input kernel for OKL is the same as the covariance function of the processes $Z_k$. Performances are measured by the reconstruction MSE (mean squared error), which is computable since we have also access to the generated non-corrupted signals. Figure \ref{fig:LowRankGP} reports the MSE (left panel) and the computation time needed to train OKL over a whole range of values of the regularization parameter $\lambda$ (right panel), both as a function of the parameter $p$. Observe that the training time is roughly increasing with $p$ since matrices of higher dimension are involved in the computation.

The best performance (in terms of MSE) is observed in correspondence with an intermediate value of the rank bound parameter (the vertical line corresponds to $p = 32$). Hence, the best rank is considerably lower than the “true rank” of the model used to generate the non-corrupted data (namely 50). A mismatch between the two is to be expected for finite and noisy samples. Conditions under which the rank of the original model can be recovered will depend on the sample size, the noise level, and the criteria used to choose the regularization parameter. It would be interesting to determine such conditions, a problem that we leave to future investigations.

\subsection{Pharmacological data} \label{sec04-02}

Reconstructing response curves in multiple subjects from sparse sets of individual measurements is a typical problem in pharmacology. The curves are generally expected to exhibit similar shapes but, on the same time, considerable inter-subject variability.


The \texttt{Study 810} dataset \cite{Merlo-Pich08, Gomeni09} has been obtained from a multicentric clinical trial (Study 810) for testing the efficacy of paroxetine (an antidepressant drug). The dataset contains time profiles of the so-called Hamilton Depression Rating Scale (HAMD) score for several patients suffering from major depressive disorders. The HAMD is an index obtained by processing a multiple choice questionnaire, that is used to measure the severity of a patient's major depression. The patients under study were either treated with placebo or administered paroxetine at two different doses. The HAMD score was evaluated at several visits, planned for each patient at weeks 0,1,2,3,4,6,and 8. The drug is considered effective if a significant decreasing in the HAMD over the weeks is observed in the patients under treatment. Due to frequent dropouts (patients abandoning the study before its completion), several of the scores are missing, especially in the last weeks. Patients with missing scores cannot be simply removed from the dataset, since dropouts are highly correlated with non-decreasing of the HAMD score, and their removal would bias the efficacy assessment. Reconstructing the missing scores is a multi-task learning problem where each patient corresponds to a task, inputs are the time instants, and outputs are the scores. A full-rank ($p=m$) weighted OKL method has been applied, by modeling the HAMD score time profiles as piece-wise linear functions. This can be done by simply using a linear spline input kernel:
\[
K(t_1,t_2) = 1 + \min\{t_1,t_2\}.
\]

The overall dataset contains 2855 scores for 494 patients. Following the setup of \cite{Dinuzzo11c}, we extracted a test set containing 1012 scores, including all the scores taken after the third week for a subset of 450 randomly chosen patients. The remaining 1843 examples are further divided into a validation set containing 553 (about $30\%$) examples, and a training set containing the remaining 1290 examples (about $70\%$). The random training/validation selection is repeated 50 times, and each time the RMSE on both the validation and the test set is computed. The regularization parameter is chosen so as to minimize the average validation error. Table \ref{TAB01} reports the average and standard deviation of the test RMSE obtained by using OKL, a nuclear norm regularized low-rank matrix approximation method (RMF), the pooled, and the independent baselines, showing a clear advantage of the OKL multi-task approach. The numerical rank of the best found output kernel is 7. Training over a whole regularization path took about 36 seconds in the average.

\begin{table}[h]
\caption{\texttt{Study 810} dataset: best average RMSE on test data (and their standard deviation over the 50 splits)}
\label{TAB01}
\begin{center}
\begin{tabular}{|c|c|c|c|}
  \hline
  Pooled   & Independent    &  RMF & OKL   \\
  \hline
  6.86 (0.02)  &    6.72(0.16)   &  6.66(0.4) & \textbf{5.37}(0.2)    \\
  \hline
\end{tabular}
\end{center}
\end{table}

\begin{figure}
  \centerline{\includegraphics[width=0.6 \columnwidth]{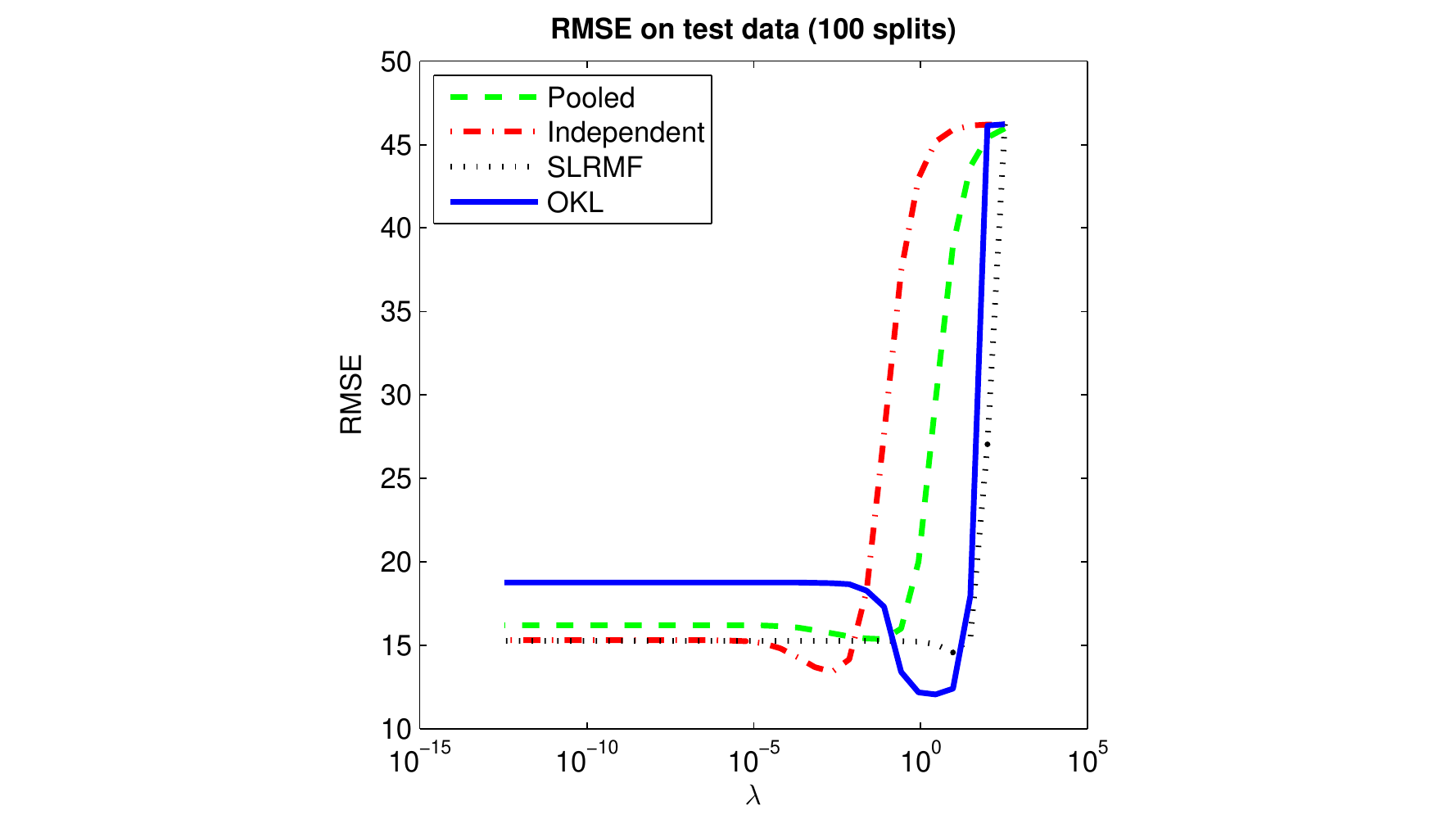}}
  \caption{\texttt{PK-PD 27} dataset: average RMSE on the test data as a function of the regularization parameter $\lambda$.}\label{FIG01}
\end{figure}

\begin{figure}
  \centerline{\includegraphics[width=0.8 \columnwidth]{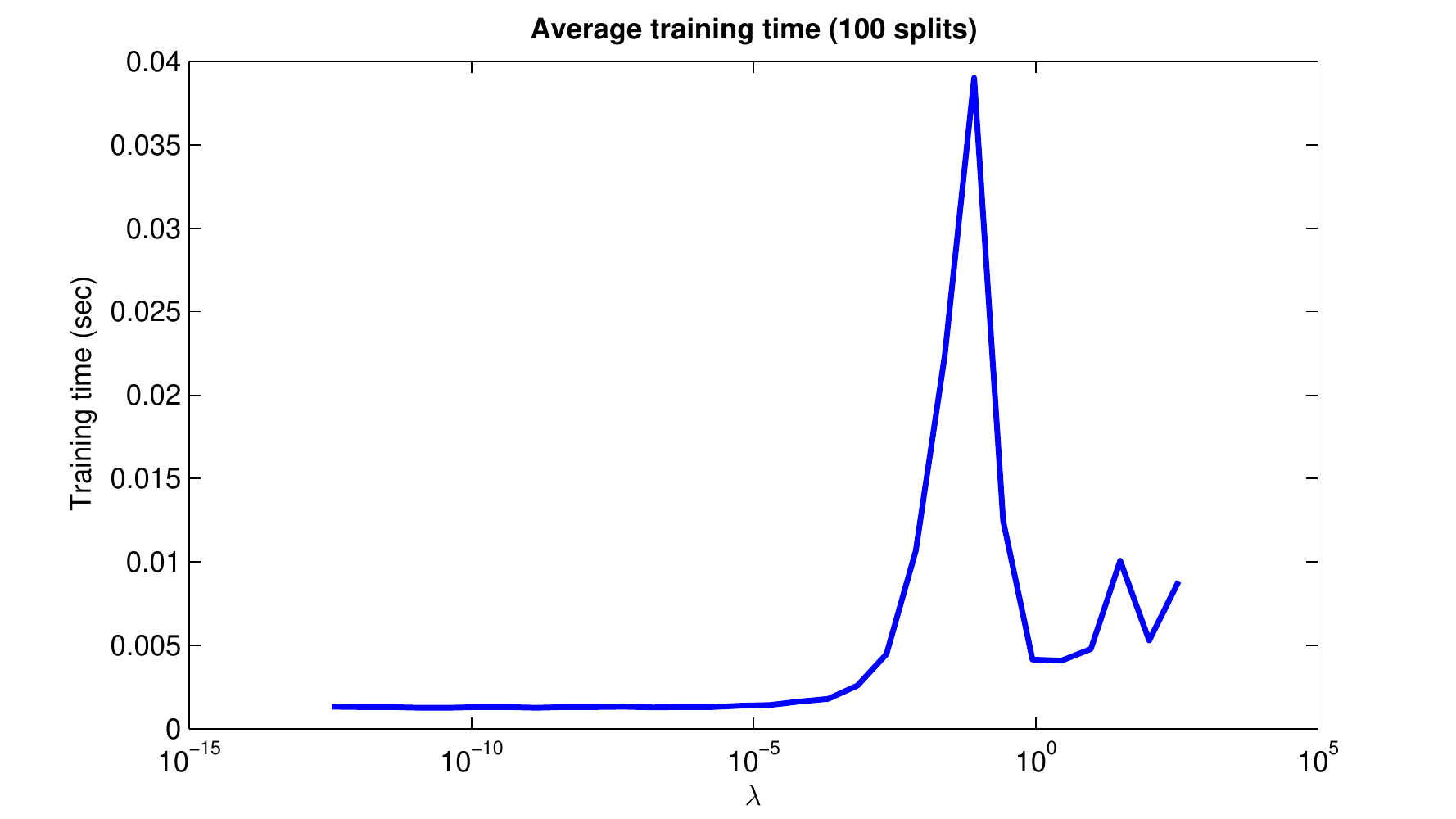}}
  \caption{\texttt{PK-PD 27} dataset: average training time of OKL as a function of the regularization parameter $\lambda$. The total training time over all the values of $\lambda$ (area under the curve) is about 1.5 seconds.}\label{FIG02}
\end{figure}

The \texttt{PK-PD 27} dataset \cite{Rocchetti97, Neve07} contains xenobiotics concentration time profiles for 27 human subjects, with samples taken at $\{0.5, 1, 1.5, 2, 4, 7, 12, 24\}$ hours after a drug administration. The goal is to reconstruct the continuous-time response curves over the non-negative real time semiline, a multi-task learning problem where the tasks are the concentration time profiles for each subject, and the inputs are the time instants. We apply the weighted OKL method described in this paper with full rank ($p=m$), as well as RMF, the independent, and the pooled baselines. The input kernel is chosen as
\[
K(t_1,t_2) = t_1 t_2 W(h(t_1),h(t_2)),
\]
\noindent where $W$ is the cubic spline kernel
\[
W(x_1,x_2) = \frac{x_1 x_2 \min\{x_1,x_2\}}{2}-\frac{(\min\{x_1,x_2\})^3}{3},
\]
\noindent and $h$ is a simple transformation designed to obtain an asymptotic decay to zero of the response curve:
\[
h(t) = (1+t)^{-1}.
\]
The kernel is designed so as to incorporate the available prior knowledge about the typical shape of a concentration response curve, which has to be smooth, with zero initial condition, and asymptotically decaying to zero. In order to simulate a realistic sparse sampling scenario, we follow the approach of \cite{Pillonetto10}, where only 3 measurements per subject (out of the 8 available) are randomly selected for training. The remaining samples are used for test. The selection is repeated 100 times, and the results are averaged. Table \ref{TAB02} reports the values of average RMSE in correspondence with the best value of $\lambda$, again showing an advantage of OKL over the other methods. Figure \ref{FIG01} reports the average RMSE as a function of the regularization parameter for the different method. Figure \ref{FIG02} reports the average training time of OKL as a function of the regularization parameter. The numerical rank of the best found output kernel is 27 (therefore, in this case we get a full rank solution). Training OKL over a whole regularization path took about 1.5 seconds in the average.

\begin{table}[h]
\begin{center}
\caption{\texttt{PK-PD 27} dataset: best average RMSE on test data (and their standard deviation over the 100 splits)}
\label{TAB02}
\begin{tabular}{|c|c|c|c|}
  \hline
  Pooled   & Indep.    &  RMF & OKL   \\
  \hline
  15.4(0.68)   &  13.4(1.55)   &  14.6(1.97) & \textbf{12} (0.92)       \\
  \hline
\end{tabular}
\end{center}
\end{table}

\subsection{Collaborative filtering (MovieLens) data} \label{sec04-03}

The MovieLens datasets \footnote{\url{http://www.grouplens.org/}} are popular collaborative filtering benchmarks, containing collections of ratings in the range $\{1, \ldots, 5\}$ assigned by several users to a set of movies. The goal is to learn the preferences of each user for all the movies. This can be interpreted as a multi-task learning problem, where each task is the preference function of one of the users. Currently, three datasets of different sizes are available, see Table \ref{TAB03}. In addition to the ratings, the datasets contain additional metadata associated with the movies (e.g. genre and title), the users (e.g. gender, age, occupation) or the ratings themselves (timestamp, tags).

\begin{table}[h]
\caption{\texttt{MovieLens} datasets: total number of users, movies, and ratings.} \label{TAB03}
\begin{center}
\begin{tabular}{|c|c|c|c|}
    \hline
    Dataset         & Users  & Movies & Ratings \\
    \hline
    \texttt{MovieLens100K}   & 943   & 1682  & $10^5$ \\
    \texttt{MovieLens1M}     & 6040  & 3706  & $10^6$ \\
    \texttt{MovieLens10M}    & 69878 & 10677 & $10^7$ \\
  \hline
\end{tabular}
\end{center}
\end{table}

The weighted OKL method described in this paper has been applied to the three MovieLens datasets. The input kernel uses the movie's metadata, while the similarity between the users is learned automatically from the data. Although the framework allows to incorporate any type of movie metadata, only the Id and the genre (present in all three datasets) have been used for simplicity. The input kernel has been designed as
\[
K(x_1,x_2) = \delta_K(x_1^{id},x_2^{id}) + \textrm{exp}\left(-d_H(x_1^g,x_2^g)\right),
\]
\noindent where $\delta_K$ is the Kronecker delta kernel (non-zero only when the movie Ids are equal), $x_1^g, x_2^g$ are binary vectors encoding the genre metadata, and $d_H$ is the normalized Hamming distance. It can be easily shown that $K$ is a valid positive semidefinite kernel. In these problems, keeping the $\delta_K$ kernel part is important, since it allows to treat two distinct movies with identical metadata as different.

Learning performance are evaluated using both the root mean squared error (RMSE), and the normalized mean absolute error (NMAE). The latter is obtained by normalizing the mean absolute error (MAE) by a factor that depends on the range of the ratings, that is $\textup{NMAE} = \textup{MAE}/(r_{\textup{max}}-r_{\textup{min}})$, see e.g. \cite{Goldberg01}. For the MovieLens datasets, we have $r_{\textup{min}} = 1$ and $r_{\textup{max}} = 5$, so that $\textup{NMAE} = 0.25 \cdot \textup{MAE}$. For \texttt{MovieLens100K} and \texttt{MovieLens10M}, performance are evaluated on the test sets $r_a$ and $r_b$ provided with the datasets. The dataset \texttt{MovieLens1M} does not come with predefined test sets, therefore we also extracted a random test set containing about the $50\%$ of the ratings for each user, a setup adopted in \cite{Toh09, Jaggi10}. The regularization parameter is tuned automatically by minimizing over a broad range the performance measure (RMSE or NMAE) on a validation set containing $25\%$ of the non-test examples for each user. Only the remaining $75\%$ are used for training. The results are summarized in Table \ref{TAB04}. In all the experiments, we use raw data without any normalization, and the rank of the output kernel has been limited to $p =5$. The rank constraint turned out to be active for all the optimal kernels. The multi-task OKL method systematically outperforms RMF, as well as the two single-task baselines. Other recent results on these datasets under various experimental settings can be found, for example, in \cite{Rennie05, DeCoste06, Abernethy09, Toh09, Lawrence09, Jaggi10}.

\begin{table}[h]
\caption{\texttt{MovieLens} datasets: test RMSE (above) and NMAE (below) for different dataset splittings for weighted OKL, RMF, the pooled and the independent baselines.}
\label{TAB04}
\begin{center}
\begin{tabular}{|c|c|c|c|c|}
  \hline
   Test         & Pooled    & Independent       & RMF     & OKL \\
  \hline

  \hline
  \multicolumn{5}{c}{\texttt{MovieLens100K} }\\
  \hline

  \multirow{2}{*}{$r_a$}        &  1.0371   &  1.0605           &  1.0007               &\textbf{0.9751}   \\
                                &  0.2087   &  0.2147           &  0.1949               &\textbf{0.1906}   \\
  \hline
  \multirow{2}{*}{$r_b$}        &  1.0423   &  1.0809           &  1.0296               &\textbf{0.9958}    \\
                                &  0.2099   &  0.2181           &  0.2019               &\textbf{0.1942}    \\
  \hline
  \multirow{2}{*}{$50\%$}       &  1.0209   &  1.0445           &  1.0300               &\textbf{0.9557}    \\
                                &  0.2043   &  0.2109           &  0.1993               &\textbf{0.1893}    \\
  \hline
  \multicolumn{5}{c}{\texttt{MovieLens1M} }\\
  \hline
  \multirow{2}{*}{$50\%$}       & 0.9811    &   1.0297           &  0.9023          &\textbf{0.8945}    \\
                                & 0.1961    &   0.2070           &  0.1761          &\textbf{0.1752}    \\
  \hline
  \multicolumn{5}{c}{\texttt{MovieLens10M} }\\
  \hline
  \multirow{2}{*}{$r_a$}          &  0.9989   &     1.0344      &   1.6501      &   \textbf{0.9427}    \\
                                  &  0.1970   &     0.2063      &   0.2892      &   \textbf{0.1806}    \\
  \hline
  \multirow{2}{*}{$r_b$}                &  0.9810   &    1.0211         &      0.9296       &\textbf{0.9141}    \\
                                        &  0.1926   &    0.2034         &      0.1806       &\textbf{0.1756}   \\
  \hline
  \multirow{2}{*}{$50\%$}               &  0.9441   &    0.9721         &       0.8627      &\textbf{0.8501}    \\
                                        &  0.1846   &    0.1915         &       0.1642      &\textbf{0.1624}    \\
  \hline
\end{tabular}
\end{center}
\end{table}

\section{Conclusions and future developments} \label{sec05}

Learning multiple tasks and simultaneously inferring the relationships between them is possible by using a kernel-based method that learns a decomposable multi-task kernel from multiple datasets. By employing a block coordinate descent strategy and iterative solvers for linear operator equations, it is possible to efficiently obtain a minimizer that yields good predictive performances. The method obviates the issue of manually specifying the similarities between the tasks. In addition, a systematic learning performance improvement with respect to single-task baselines and standard regularized low-rank matrix approximation can be observed on several datasets.

In the future, it would be worthwhile to extend the proposed method by using loss functions different from the the least square loss. Also, it would be interesting to extend the framework so as to exploit other types of structural knowledge about the task relationships, for instance along the lines of \cite{Argyriou08, Romera12}.

\appendix

\section{Proofs}

\noindent \textbf{Proof of Lemma \ref{LEM01}}

Any optimal  $\mathbf{A} \in \mathbb{R}^{\ell \times p}$ for problem (\ref{EQ04}) admits a unique decomposition of the form
\[
\mathbf{A} = \mathbf{C}\mathbf{B} + \mathbf{U}, \qquad \mathbf{U}\mathbf{B}^T = \mathbf{0}.
\]
\noindent By letting $\Ky = \mathbf{B}\mathbf{B}^T$, it follows that $\Ky \in \mathbb{S}^{m,p}_+$, and we have
\[
\Kx\mathbf{A}\mathbf{B}^T = \Kx\mathbf{C}\mathbf{B}\mathbf{B}^T = \Kx\mathbf{C}\Ky.
\]
\noindent In addition, we have
\begin{align*}
\frac{\dotprod{\mathbf{A}}{\Kx\mathbf{A}}_{F}}{2} & = \frac{\dotprod{\mathbf{U}}{\Kx\mathbf{U}}_{F}}{2} + \frac{\dotprod{\mathbf{C}^T\Kx\mathbf{C}}{\Ky}_{F}}{2}\\
 & \geq \frac{\dotprod{\mathbf{C}^T\Kx\mathbf{C}}{\Ky}_{F}}{2}.
\end{align*}
\noindent It follows that we can set $\mathbf{U} = \mathbf{0}$ and $\mathbf{A} = \mathbf{C}\mathbf{B}$, without any loss of generality. 
\begin{flushright}
    $\Box$
\end{flushright}

\noindent  \textbf{Proof of Lemma \ref{LEM02}}

Letting $\mathbf{\Theta} = \mathbf{C} \Ky$, problem (\ref{EQ02}) can be rewritten as
\begin{align*}
& \min_{\mathbf{\Theta} \in \mathbb{R}^{\ell \times m}}\left(\frac{\|\mathbf{Y}-\Kx\mathbf{\Theta}\|_{\mathbf{W}}^2}{2\lambda}+ \min_{\Ky \in \mathbb{S}^{m,p}_+} N_{\mathbf{\Theta}}(\Ky)\right),\\
& \st \quad \textrm{rg}(\mathbf{\Theta}) \subseteq \textrm{rg}(\Ky),
\end{align*}
\noindent where
\[
N_{\mathbf{\Theta}}(\Ky) = \frac{\langle \mathbf{\Theta}^T\Kx\mathbf{\Theta}, \Ky^{\dag} \rangle_F }{2}+ \frac{\textrm{tr}(\Ky)}{2}
\]
\noindent A minimizer of $N_{\mathbf{\Theta}}(\Ky)$ can be expressed in closed-form:
\[
\Ky = \left(\mathbf{\Theta}^T\Kx\mathbf{\Theta}\right)^{1/2},
\]
\noindent so that
\[
\min_{\Ky \in \mathbb{S}^{m,p}_+} N_{\mathbf{\Theta}}(\Ky) = \textup{tr}\left(\left(\mathbf{\Theta}^T\Kx\mathbf{\Theta}\right)^{1/2}\right).
\]
\noindent All the constraints are satisfied provided that
\[
\textrm{rank}(\mathbf{\Theta}) = \textrm{rank}(\mathbf{\Ky}) \leq p.
\]
\begin{flushright}
    $\Box$
\end{flushright}

\section{Matrix identities}

We recall some identities involving the vectorization operator, the Kronecker and the Hadamard products, see e.g. \cite{Horn91}:
\begin{equation}\label{EQ08}
\textrm{vec}\left(\mathbf{A}\mathbf{X}\mathbf{B}\right) = \left( \mathbf{B}^T \otimes \mathbf{A} \right) \textrm{vec}(\mathbf{X}),
\end{equation}
\begin{equation}\label{EQ09}
\left(\mathbf{A}\otimes \mathbf{B}\right)\left(\mathbf{C}\otimes \mathbf{D}\right) = \left(\mathbf{A}\mathbf{C}\right)\otimes \left(\mathbf{B}\mathbf{D}\right),
\end{equation}
\begin{equation}\label{EQ10}
\textrm{vec}\left(\mathbf{A} \odot \mathbf{B}\right) = \textrm{diag}(\textrm{vec}\left(\mathbf{A}\right))  \textrm{vec}\left(\mathbf{B}\right).
\end{equation}
These identities hold whenever the sizes of the involved matrices are compatible.

\bibliographystyle{plain}







\end{document}